\documentclass[journal]{IEEEtran}
\ifCLASSINFOpdf
\else
\fi
\usepackage[cmex10]{amsmath}
\usepackage{array}
\usepackage{mdwmath}
\usepackage{mdwtab}
\usepackage{eqparbox}
\usepackage{url}
\usepackage{textcomp}
\usepackage{mathtools}
\usepackage[thinc]{esdiff}
\usepackage{hyperref}
\usepackage{booktabs} 
\usepackage{subfigure}
\usepackage{graphicx}
\usepackage{tabularx}
\usepackage{cite}

\begin{document}
%
\title{A Reconfigurable Pneumatic Joint Enabling\\ Localized Selective Stiffening and Shape Locking\\in Vine-Inspired Robots}
%
%
%

\author{Ayodele~James~Oyejide,~\IEEEmembership{Student Member,~IEEE,}
Ustaz~A.~Yaqub,
Samir~Erturk,
Eray~A.~Baran,~\IEEEmembership{Member,~IEEE,}
Fabio~Stroppa,~\IEEEmembership{Senior Member,~IEEE}%
\thanks{Ayodele J. Oyejide is with the Department of Electrical and Electronics Engineering, Kadir Has University, Cibali, Istanbul, Türkiye (e-mail: aoyejide@stu.khas.edu.tr).}%
\thanks{Ustaz A. Yaqub, Samir Ertürk, and Fabio Stroppa are with the Department of Computer Engineering, Kadir Has University, Cibali, Istanbul, Türkiye.}%
\thanks{Eray A. Baran is with the Department of Mechatronics Engineering, Istanbul Bilgi University, Istanbul, Türkiye (e-mail: eray.baran@bilgi.edu.tr).}%
\thanks{Manuscript received April XX, 2026; revised XXXXXXXXX XX, 2026.}}

%
%


\markboth{Oyejide \textit{et al}}%
{Shell \MakeLowercase{\textit{Oyejide et al.}}: Bare Demo of IEEEtran.cls for Journals}
%

 



\maketitle


\begin{abstract}
Vine-inspired robots achieve large workspace coverage through tip eversion, enabling safe navigation in confined and cluttered environments. However, their deployment in free space is fundamentally limited by low axial stiffness, poor load-bearing capacity, and the inability to retain shape during and after steering. In this work, we propose a reconfigurable pneumatic joint (RPJ) architecture that introduces discrete, pressure-tunable stiffness along the robot body without compromising continuous growth. Each RPJ module comprises symmetrically distributed pneumatic chambers that locally increase bending stiffness when pressurized, enabling decoupling between global compliance and localized rigidity. We integrate the RPJs into a soft growing robot with tendon-driven steering and develop a compact base station for mid-air eversion. System characterization and experimental validation demonstrate moderate pressure requirements for eversion, as well as comparable localized stiffening and steering performance to layer-jamming mechanisms. Demonstrations further show that the proposed robot achieves improved shape retention during bending, reduced gravitational deflection under load, cascading retraction, and reliable payload transport up to $202$~g in free space. The RPJ mechanism establishes a practical pathway toward structurally adaptive vine robots for manipulation-oriented tasks such as object sorting and adaptive exploration in unconstrained environments.
\end{abstract}


%
\IEEEpeerreviewmaketitle

\section{Introduction}
\label{sec:introduction}

\IEEEPARstart{V}{ine}-inspired robots, modeled after the growth mechanics of climbing plants, have recently emerged as a promising class of continuum soft robots. Unlike robots that rely on body articulation or wheel-ground interaction for movement, vine robots achieve locomotion through \textit{tip eversion}, growing from the tip rather than translating their body. This enables continuous elongation with minimal frictional interaction between the robot body and the surrounding environment~\cite{hawkes2017soft}. This tip-eversion capability has enabled vine robots to be deployed in diverse applications, including archaeological exploration and operation within fragile ruins~\cite{coad2019vine, der2021roboa}, coral reef inspection~\cite{luong2019eversion}, odor sensing~\cite{oyejide2026softgrowing}, and clinical demonstrations~\cite{takahashi2022inflated, oyejide2025miniaturized, berthet2021mammobot}.

\begin{figure}[t!]
    \centering
    \includegraphics[width=\linewidth]{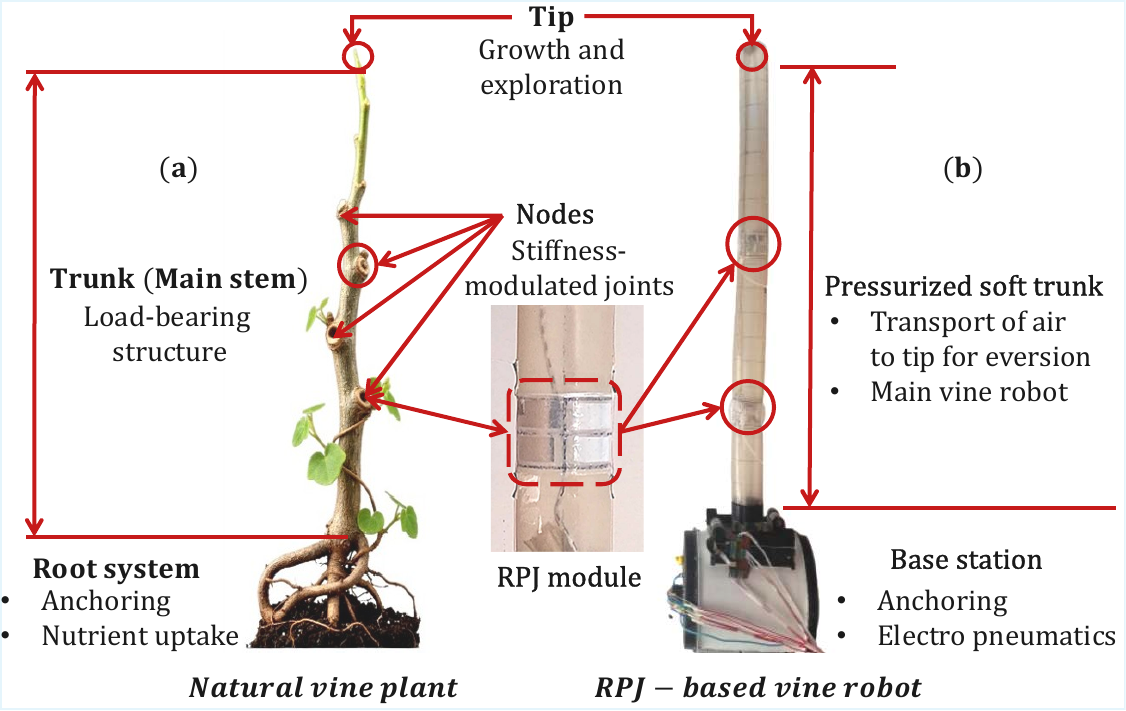}
    \caption{Proposed concept for achieving localized stiffness modulation and shape locking in vine robots. (a) Biological vine structure, consisting of the root, trunk, nodes, and tip. (b) Bioinspired abstraction, where the nodal element is translated into a soft reconfigurable pneumatic joint (RPJ) for localized stiffness control. Detailed discussion is provided in Sec.~\ref{subsec:biomimicry}}.
    \label{fig:RPJ_based_vinerobot_overview}
\end{figure}

However, vine robots inherently possess highly compliant bodies. As a result, achieving active localized stiffness required for navigation, load-bearing, and shape locking without sacrificing global softness remains challenging.

Early studies relied on passive shape locking, where the robot grows along or into surrounding structures and adopts the shape of the environment without active stiffness control~\cite{hawkes2017soft, blumenschein2020design}. While effective in certain scenarios, this approach restricts the robot to applications that rely primarily on passive navigation. Alternatively, shape locking can be achieved through preformed bending~\cite{blumenschein2021geometric}. However, this results in largely unreconfigurable structures that limit localized bending and adaptability.

To address these limitations, several stiffness-modulation strategies have been proposed to enable localized reinforcement and shape control in vine robots. These include passive locking mechanisms~\cite{jitosho2023passive, hawkes2017soft}, layer-jamming techniques for tunable rigidity~\cite{do2020dynamically, do2024stiffness, li2025enhanced, li2025mixedlayerjamming, exarchos2022task}, mechanical locking and geometric constraint mechanisms~\cite{jitosho2023passive, li2021bioinspired, fuentes2023deployable}, pneumatic artificial muscles~\cite{kubler2024comparison, kubler2023multi, hawkes2017soft, feteih2025active, du2023finite}, and thermally responsive materials such as phase-change alloys~\cite{al2025phase} and actuating polymers~\cite{krishna2025towards}. A review of these approaches, including their operating principles and limitations, is provided in Section~\ref{sec:related_works}.

Despite these developments, most existing stiffness-modulation and shape-locking strategies introduce significant mechanical or fabrication complexity and often compromise the intrinsic global compliance of vine robots~\cite{li2025enhanced, jitosho2023passive, blumenschein2021geometric}. More importantly, existing approaches either do not enable truly selective localized stiffness along the robot body~\cite{jitosho2023passive, blumenschein2021geometric, al2024variable} or limit the robot’s ability to grow and retract freely from the base~\cite{jitosho2023passive}. The ability to selectively introduce localized stiffness while preserving global flexibility and eversion cycles is therefore critical for improving the dexterity and functionality of vine robots.

In this work, we propose a reconfigurable pneumatic joint (RPJ) architecture for vine-inspired robots that bridges the compliance-stiffness gap in existing designs. The RPJ system introduces pneumatically actuated nodes embedded between consecutive vine robot segments, each capable of independently controlling stiffness and curvature. The concept draws inspiration from the segmented articulation observed at plant nodes, particularly the vine, which reinforce the structure and allow the vine plant to support leaves and fruit clusters (Fig.~\ref{fig:RPJ_based_vinerobot_overview}). We reinterpreted this biological principle using compliant materials, specifically low-density polyethylene (LDPE) and extensible fabrics, to realize a soft, reconfigurable pneumatic mechanism for selective stiffening. 

While existing approaches enable localized stiffening to varying degrees, the proposed RPJ mechanism also preserves the intrinsic flexibility of the vine robot body while introducing new capabilities of: \textit{(i)} predictable shape locking, \textit{(ii)} unique spatial morphologies, and \textit{(iii)} cascading retraction, which, to the best of our knowledge, has not been demonstrated in previous studies.

The key contributions of this work are as follows:
\begin{enumerate}
\item the RPJ mechanism that provides localized stiffness, curvature control, load-bearing capacity, and shape locking in vine robots while preserving global flexibility;
\item a combined analytical and experimental characterization of RPJ behavior, establishing quantitative relationships between chamber pressure, contact force distribution, and effective stiffness for design and control;
\item a reconfigurable vine robot system that enables sustained operation in free space, where the structure reconfigures under its own weight without reliance on environmental support; and
\item an RPJ-enabled cascading retraction strategy that allows controlled segment-wise inversion of the robot body without additional onboard retraction mechanisms.
\end{enumerate}

\section{Related Works}
\label{sec:related_works}

In this section, we review existing stiffness modulation and shape locking strategies used in vine robots. We discuss their operating principles, advantages, and drawbacks.

\subsection{Pneumatic Artificial Muscles}
\label{subsec:pneumatic_artificial_muscles}

Pneumatic artificial muscles (PAMs) are soft actuators that generate tensile contraction when pressurized. They represent one of the earliest steering and stiffening mechanisms used in vine robot research~\cite{hawkes2017soft}. These actuators consist of pressurizable chambers that constrain radial expansion, producing tensile forces capable of bending or steering the robot body~\cite{kubler2024comparison}. Several variants have been implemented in vine robots, including series pneumatic artificial muscles (sPAMs)~\cite{hawkes2017soft, blumenschein2020design}, series pouch motors (SPMs)~\cite{coad2019vine}, cylindrical pneumatic artificial muscles (cPAMs), and fabric pneumatic artificial muscles (fPAMs)~\cite{kubler2024comparison}.

PAMs are typically mounted along the outer surface of the inflatable trunk~\cite{kubler2024comparison} or embedded within the robot structure~\cite{feteih2025active}. When one or more muscles are actuated, contraction on the corresponding side generates differential strain across the trunk, producing curvature toward the actuated side and enabling distributed steering during navigation. Consequently, several studies have exploited PAM contraction forces to influence the mechanical stiffness of the robot body~\cite{hawkes2017soft, feteih2025active}. Demonstrations include steering in everting vine robots of $1$–$2$ m using sPAMs and $7$–$10$ m using SPMs~\cite{coad2019vine, blumenschein2020design}.

However, this stiffness modulation arises indirectly through tension-induced reinforcement rather than a dedicated locking mechanism. Because PAMs typically span large portions of the trunk, the resulting stiffening effect is distributed along the actuator length, limiting the ability to localize stiffness changes to specific robot segments. Moreover, these actuator designs introduce additional limitations. Repeated pressurization can produce stress concentrations that accelerate actuator fatigue, while long fluidic pathways introduce internal resistance that can delay actuation of distal segments in long robots~\cite{blumenschein2020design}.

\subsection{Layer Jamming-based Stiffening}
\label{subsec:layer_jamming}

In vine-inspired robots, layer jamming mechanisms are typically implemented by embedding stacks of thin sheets or fabrics within the robot body or by attaching laminated sleeves along selected trunk segments~\cite{do2024stiffness, li2025enhanced, li2025mixedlayerjamming}. When vacuum is applied, or internal pressure is evacuated~\cite{do2020dynamically}, the stacked layers jam together, increasing the effective bending stiffness of the robot along the jammed region. This approach enables reversible stiffness modulation without rigid mechanical components, making it attractive for soft robotic systems that must maintain compliance during navigation.

Despite its effectiveness, layer jamming inherently produces stiffening over a continuous region rather than at discrete structural nodes~\cite{do2024stiffness}. Because the stacked layers extend along finite trunk sections, the jamming process increases bending resistance across the entire layered segment. Consequently, the transition between compliant and stiff regions becomes distributed rather than sharply localized~\cite{exarchos2022task, feteih2025active}. This distributed stiffening reduces the robot's ability to achieve large curvature changes required for confined maneuvering.

Although stiffness modulation can be achieved by selectively jamming discrete sections, the approach is constrained by the jamming material reaching a maximum bulk stiffness at higher vacuum pressures~\cite{li2025mixedlayerjamming}. In addition, layered stacks introduce extra material thickness and friction interfaces that can reduce growth speed and sometimes hinder retraction. Recent layer-jamming implementations also require specialized fabrication strategies, increasing system complexity and integration demands~\cite{li2025enhanced, li2025mixedlayerjamming}.

\subsection{Mechanical Locking Mechanisms}
\label{subsec:mechanical_loacking}

Mechanical locking mechanisms provide variable stiffness by introducing discrete structural constraints along the robot body. Unlike friction-based or pressure-induced approaches (Section~\ref{subsec:layer_jamming} and Section~\ref{subsec:pneumatic_artificial_muscles}), these systems rely on physical engagement between structural elements that restrict relative motion between adjacent segments. For example, a study implemented a fastener-hook locking mechanism~\cite{jitosho2023passive}. Although this approach enabled selective segment stiffening to create multi-bending configurations, it achieved a relatively low stiffening ratio of $1.3\times$ and required additional locking components separate from the actuation system.

Other approaches include pressure-triggered latching mechanisms that enable multi-segment steering~\cite{hawkes2017soft}, as well as a steering-reeling mechanism (SRM). The SRM induces controlled buckling through an internal rigid steering structure, generating large bending angles at discrete locations along the robot length~\cite{haggerty2021hybrid}. However, latch-based systems can be difficult to manufacture and may exhibit limited reversibility. Similarly, the SRM introduces rigid internal components that reduce growth speed and limit the robot’s ability to traverse gaps smaller than the SRM diameter.

Magnetically assisted couplings have also been proposed as locking mechanisms~\cite{kubler2023multi, watson2020permanent}. These components are integrated as discrete modules along the trunk, where engagement between magnetic elements increases local bending stiffness by preventing relative motion. While mechanical locking can provide substantial stiffness through geometric engagement, the required rigid interfaces introduce structural discontinuities along the otherwise soft robot body. Consequently, locked sections behave as rigid links connected by compliant segments, altering the deformation behavior of the surrounding structure. In addition, reliable engagement and disengagement require precise alignment, increasing mechanical complexity and integration requirements.

\subsection{Low Melting Point Alloy and Actuating Polymer Stiffening}

Thermally responsive materials represent another approach to stiffness modulation in vine robots, although they have been used relatively infrequently. Phase-change alloy~\cite{al2025phase}, low-melting-point alloys~\cite{al2024variable}, and thermally activated polymers~\cite{krishna2025towards} modify their mechanical properties in response to temperature changes, enabling transitions between compliant and stiff states. At elevated temperatures the material softens, while cooling restores structural rigidity. Hence, these materials can act as controllable reinforcement elements whose stiffness depends on the applied thermal input~\cite{krishna2025towards}.

For example, low-melting-point alloys have been demonstrated as both a stiffening medium and a working fluid for growth in a $6$ mm vine robot~\cite{al2024variable}. Although this method achieved a stiffening ratio of $20\times$, the robot cannot grow while stiffened because the medium fully solidifies, and phase-change transitions between soft and solid states are relatively slow. These characteristics make repeatable bending difficult and introduce challenges associated with thermal management and electrical routing, increasing overall system complexity.

\begin{table}[t]
\centering
\caption{Variables used in the RPJ mechanical model}
\label{tab:RPJ_variables}
\begin{tabular}{ll}
\hline
Symbol & Description and Units \\
\hline
$p_j$ & Internal pressure of RPJ chambers (Pa) \\
$L_j$ & Axial length of an RPJ module (m) \\
$L_t$ & Length of the trunk segment excluding the RPJ (m) \\
$w_c$ & Unpressurized chamber width (m) \\
$h_c$ & Unpressurized chamber height (m) \\
$r_c$ & Effective chamber radius (m) \\
$A_c$ & Effective contact area of a single chamber (m$^2$) \\
$E$ & Elastic modulus of LDPE (Pa) \\
$(EI)_{\text{\textit{eff}}}$ & Effective bending stiffness (N\,m$^2$) \\
$\kappa$ & Curvature (m$^{-1}$) \\
$M$ & Bending moment (N\,m) \\
$T$ & Tendon tension (N) \\
$K$ & Effective column length factor (--) \\
\hline
\end{tabular}
\end{table}

\section{Proposed Reconfigurable Pneumatic Joint (RPJ) System}
\label{sec:design_operating_principle}
This section presents the geometric architecture, pneumatic operating principles, and mechanical modeling of the RPJ integrated within the vine robot. It further describes the coupling between pneumatic stiffening and tendon actuation for directional steering. Table~\ref{tab:RPJ_variables} summarizes the variables used in the RPJ model. The subscripts $j$ and $c$ denote quantities associated with a joint and the joint chamber, respectively.

\subsection{Biomimetic Abstraction of RPJ-based Vine Robot}
\label{subsec:biomimicry}
As described in Fig.~\ref{fig:RPJ_based_vinerobot_overview}, vine plants are composed of four main components: (i) \textit{root system}, (ii) \textit{trunk}, (iii) \textit{nodes}, and (iv) \textit{apical meristem} (\textit{growing tip}) \cite{hellman2003grapevine}. Each component performs a distinct functional role, yet their coordinated interaction enables efficient growth, adaptive morphology, and load-bearing capability. The \textit{\textbf{root system}} anchors the plant and regulates resource uptake~\cite{hellman2003grapevine, vanden2021under}. In vine robots, this role is embodied by the base station, which anchors the vine and provides pneumatic energy for eversion~\cite{blumenschein2020design}. The \textit{\textbf{trunk}}, on the other hand, serves as the main load-bearing conduit, maintaining axial stiffness to prevent collapse under self-weight or external loads~\cite{hellman2003grapevine, bai2024behavioral}. In our robotic analogue, the trunk is a thin-walled, pressurized tube, where internal pressure generates membrane stresses that provide bending resistance and support free-space extension without rigid skeletons~\cite{blumenschein2020design}.

The \textit{\textbf{Nodes}} are discrete junctions with enhanced structural support and directional influence~\cite{yamaji2014node, hellman2003grapevine}. Mechanically, they act as localized stiffness-modulated regions along the stem. This principle directly motivates the RPJ design, enabling tunable reinforcement and discrete curvature control in the vine robot. Finally, the \textit{\textbf{apical meristem}} governs elongation, environmental sensing, and adaptive steering~\cite{hawkes2017soft, blumenschein2020design}. In our robot, this role is realized by the everting tip, separating exploration (tip) from support and anchoring (nodes). This mapping provides a biomimetic foundation for the proposed RPJ geometry, placement, and actuation strategy.

\subsection{Geometric Description}
\label{subsec:geometric_description}

\begin{figure}[h!]
    \centering
    \includegraphics[width=\linewidth]{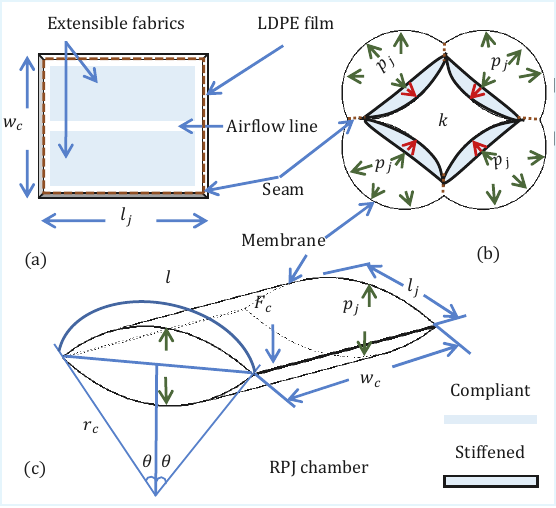}
    \caption{Schematic of an RPJ node. (a) Single fabric-reinforced chamber in the unpressurized state. (b) Circumferential cross-sectional view under uniform chamber pressure $p_j$, generating distributed contact forces (red arrows) at the joint-vine interface. These forces, aided by extensible fabric reinforcement, produce localized stiffness modulation and the resulting curvature $k$. (c) Model of the single chamber in (a) based on free pouch approximation~\cite{niiyama2015pouch}.}
    \label{fig:RPJ_geometry}
\end{figure}

The RPJ is realized as a modular chambered structure integrated inline with the tubular body of the vine robot. Fig.~\ref{fig:RPJ_geometry} illustrates the geometry of an isolated RPJ node prior to integration. Each chamber is fabricated by thermally sealing LDPE films to form a thin-walled membrane compartment of width $w_c$ and axial length $l_j$. 

The symmetric chamber arrangement is selected to ensure that pressure-induced forces are directed toward the central axis of the structure, thereby minimizing torsional deformation and enabling predictable bending under external tendon actuation. Air is supplied through a common inlet and distributed to all chambers via interconnecting flow path. In the unpressurized state, the chambers retain a seam-defined square geometry (Fig.~\ref{fig:RPJ_geometry}(a)). Upon pressurization with chamber pressure $p_j$, the membranes expand to bulging shapes constrained by the surrounding seams (Fig.~\ref{fig:RPJ_geometry}(b)). This deformation is analogous to pouch actuators~\cite{kubler2024comparison} with cylindrical surface segments, but here the seams partition the structure into discrete chambers, preventing the formation of a continuous shell (Fig.~\ref{fig:RPJ_geometry}(c)). The seam boundaries further restrict lateral expansion and preserve geometric separation between chambers, resulting in localized volume expansion within each chamber rather than global radial inflation. As a result, the RPJ node behaves as a set of constrained membrane volumes whose deformation is governed by internal pressure and seam geometry.

\begin{figure}[t!]
    \centering
    \includegraphics[width=\linewidth]{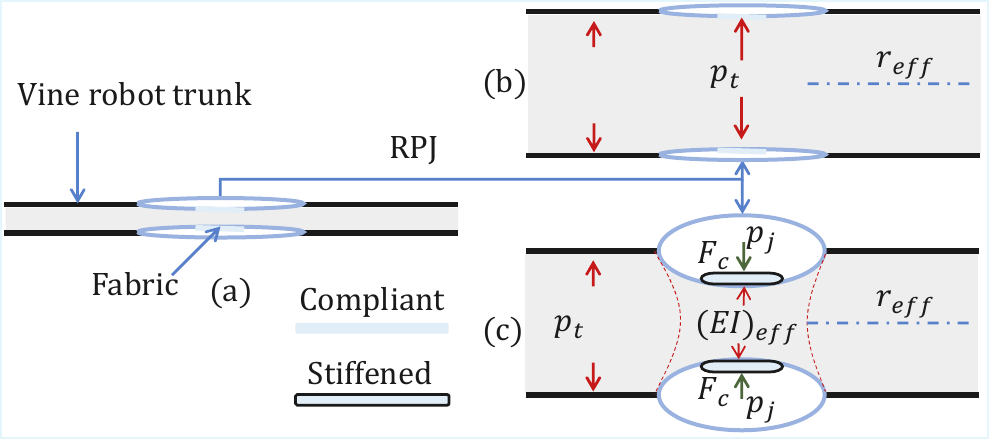}
    \caption{Interaction between the RPJ and the trunk. (a) Both RPJ and trunk unpressurized. (b) Pressurized trunk ($p_t$) with unpressurized RPJ, inducing passive stiffness at the joint due to fabric tension. (c) Pressurized RPJ ($p_j$) and trunk, where contact forces $F_c$ generated by the chambers produce active stiffening, resulting in an effective local joint stiffness $(EI)_{\mathrm{\textit{eff}}}$.}
    \label{fig:RPJ_trunk_geometry}
\end{figure}

When integrated into the vine robot, the RPJ operates under differential pressure conditions, with chamber pressure $p_j$ acting against the internal trunk pressure $p_t$ (Fig.~\ref{fig:RPJ_trunk_geometry}). To regulate this interaction, extensible laminated fabric strips are embedded near the upper and lower boundaries of each chamber (see Section~\ref{subsec:prototyping}). These reinforcements constrain excessive radial deformation and bias the resultant force toward the outer surface of the trunk, ensuring effective transmission of pressure-induced forces to the structure. 

\subsection{Chamber Deformation and Effective Stiffness}
\label{subsec:analytical_model}

The RPJ chambers are modeled as constrained pouch-like actuators whose deformation and force generation follow established pouch motor mechanics~\cite{kubler2024comparison, niiyama2015pouch}, adapted here to account for circumferential confinement and interaction with a pressurized trunk.

\subsubsection{Assumptions}
\label{subsubsec:assumption}

To obtain a tractable model of the RPJ chamber mechanics, the following assumptions are adopted: First, the chamber membrane is modeled as a thin, flexible, and inextensible film, such that deformation is governed primarily by geometric constraints rather than material stretch. Upon pressurization, each chamber inflates into a cylindrical surface segment with constant width $w_c$, while out-of-plane deformation and edge effects are neglected. We assume that the deformation is quasi-static, so inertial and dynamic effects are negligible relative to pressure-induced forces. Lastly, contact between the chamber and the trunk is continuous, and the resulting interaction force is assumed to act normal to the trunk surface.

\subsubsection{Chamber Deformation}
\label{subsubsec:chamber_deformation}

Under the stated assumptions in Section~\ref{subsubsec:assumption}, each RPJ chamber is modeled as a pressurized pouch that deforms into a cylindrical segments (Fig.~\ref{fig:RPJ_geometry}(c)). Building on the classical pouch actuator formulation~\cite{niiyama2015pouch, kubler2024comparison}, the initial flat length $l_j$, radius of curvature $r_c$, central angle $\theta$, and deformed chord length $l$ satisfy the expression in (\ref{eq:pouch_geometry}). Eliminating $r_c$ yields the normalized deformation relation (\ref{eq:pouch_contraction}), which describes the geometric contraction behavior of an individual chamber as a function of the deformation state $\theta$. It also provides a basis for selecting the initial chamber length $l_j$ to achieve a desired contraction profile under pressurization:

\begin{equation}
l_j = 2r_c\theta, \quad
r_c \sin\theta = \frac{l}{2}
\label{eq:pouch_geometry}
\end{equation}

\begin{equation}
\frac{l}{l_j} = \frac{\sin\theta}{\theta}
\label{eq:pouch_contraction}
\end{equation}

However, unlike conventional pouch actuators that produce free axial contraction, the RPJ chambers are circumferentially constrained by the central trunk. As a result, geometric contraction is inhibited, and the pressure-induced deformation is instead redirected into radial expansion. This expansion generates normal contact forces  $F_c$ at the chamber-trunk interface. Consequently, the mechanical output of the RPJ is governed not by axial shortening but by the pressure-driven interaction force, which can be expressed as a function of the deformation state $\theta$ and the pressure differential $\Delta p = p_j - p_t$.

\subsubsection{Effective Stiffness}
\label{subsubsec:stiffness_formulation}
The effective contact force generated by each chamber (governed by the pressure differential $p_j - p_t$) is expressed as in (\ref{eq:chamber_force_model}), where $A_c$ denotes the effective contact area of a single chamber:
\begin{equation}
F_c = (p_j - p_t) A_c
\label{eq:chamber_force_model}
\end{equation}

Because the chambers are symmetrically distributed around the trunk, the contact forces at the chamber-trunk interface act radially inward (Fig.~\ref{fig:RPJ_geometry}(b) and (\ref{eq:chamber_force_model})(a–c)). The lateral components cancel by symmetry, resulting in no net bending or torsional moment. Instead, the combined action of the chambers produces a circumferential constraining pressure that increases the local resistance of the trunk to deformation.

This pressure-induced interaction provides a basis for defining an effective, pressure-dependent stiffness at the joint. From (\ref{eq:chamber_force_model})(a–c), the contact force scales with the pressure differential $(p_j - p_t)$ and the chamber contact area $A_c$, acting at a radial offset $r_{\mathrm{eff}}$. These forces generate a distributed resisting moment that opposes local curvature. Accordingly, the effective bending stiffness $(EI)_{\mathrm{eff}}$ increases with $(p_j - p_t)$, $A_c$, and $r_{\mathrm{eff}}$. This formulation establishes the RPJ as a tunable stiffness element: when $p_j \approx p_t$, the joint remains compliant, whereas increasing $p_j$ enhances resistance to deformation, leading to localized stiffening without inducing torsion.

\subsection{RPJ-Tendon-Induced Bending of the Vine Robot}
\label{subsec:tendon_bending}

Directional steering is achieved via four tendons symmetrically distributed along the robot circumference. Each tendon is routed longitudinally from the base to the distal tip, and actuated through an external motorized spool at the base~\cite{exarchos2022task}. 

\begin{figure}[t!]
    \centering
    \includegraphics[width=\linewidth]{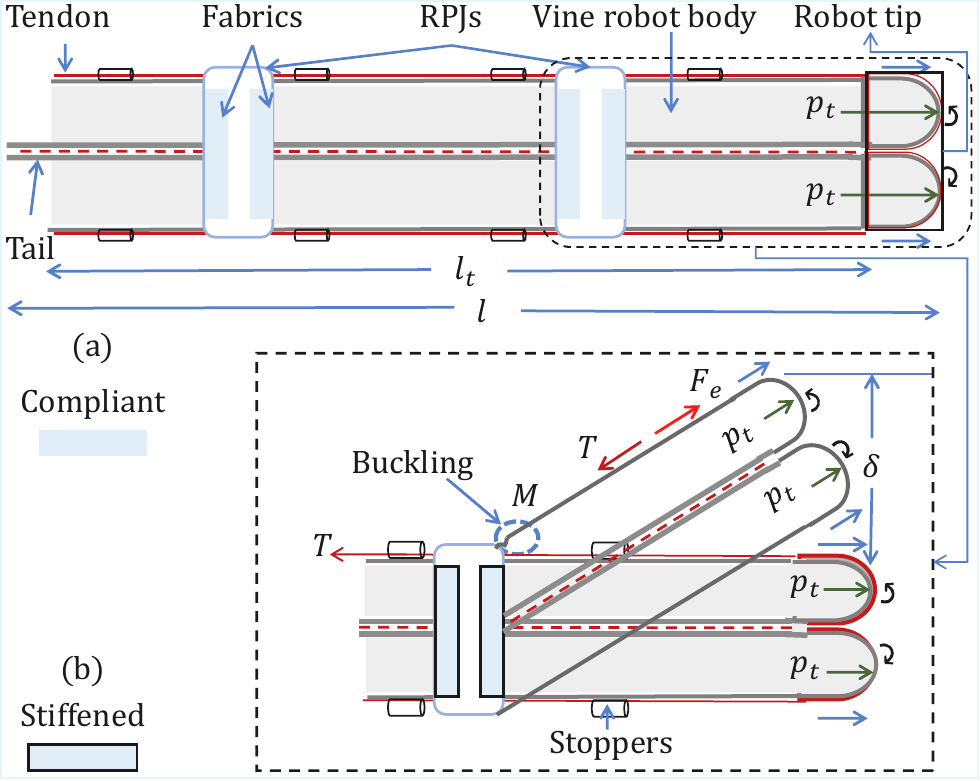}
    \caption{Side axial cross section of the RPJ-based vine robot. (a) Unactuated RPJs allow for trunk growth and showcase the tendon system. (b) Bending is initiated by stiffening the RPJs, and steering the tip by applying tension $T$ to the corresponding tendon. This tension induces buckling at the distal RPJ edge, creating a moment $M$ to lock the shape.}
    \label{fig:RPJ_trunk_steering}
\end{figure}

In the nominal configuration, all tendons are at equal length and no bending occurs. Actuation of a spool induces tendon shortening, generating a tensile force $T$ along the corresponding side of the robot, as shown in Fig.~\ref{fig:RPJ_trunk_steering}(b). The applied tension $T$ generates a bending moment about the RPJ interface as expressed in (\ref{eq:tendon_moment}), where $r$ is the radial offset of the tendon from the robot centerline: 
\begin{equation}
M = T r
\label{eq:tendon_moment}
\end{equation}
Assuming deformation is localized to the compliant segment, the resulting bending can be approximated using a beam-like relation given by (\ref{eq:tendon_curvature})~\cite{oyejide2026softgrowing}, where $\kappa$ is the curvature of the deformed segment and $(EI)_{\mathrm{eff}}$ represents the effective bending stiffness, modulated by the RPJ pressure state:
\begin{equation}
\kappa = \frac{M}{(EI)_{\mathrm{eff}}}
\label{eq:tendon_curvature}
\end{equation}

In general, steering is produced through differential tendon actuation, which introduces asymmetric tensile loading and pulls the distal segment toward the actuated side. To localize deformation, selected RPJs are pressurized to increase their bending stiffness, while the distal segment remains compliant (Fig.~\ref{fig:RPJ_trunk_steering}(a), (b)). This stiffness contrast constrains deformation upstream and concentrates bending at the interface between the stiffened RPJ and the compliant distal link. Note that during bending, not all chambers contribute equally to the deformation. Due to the directionality of tendon actuation, only the pair of chambers aligned with the bending plane are effectively engaged in an antagonistic manner. The remaining chambers have a negligible contribution to bending. This directional activation further reinforces the localization of deformation at the selected joint, establishing a decoupled actuation mechanism in which pneumatic pressure defines the spatial stiffness distribution, while tendon tension $T$ governs the direction and magnitude of bending.

Consequently, the RPJ–tendon combination enables controlled, localized rotations at discrete interfaces, allowing the robot to achieve segmented, multi-directional configurations without inducing distributed curvature along the trunk.

\subsection{RPJ-Enabled Cascading Retraction}
\label{subsec:cascading_retraction}

\begin{figure}[h!]
\centering
\includegraphics[width=\linewidth]{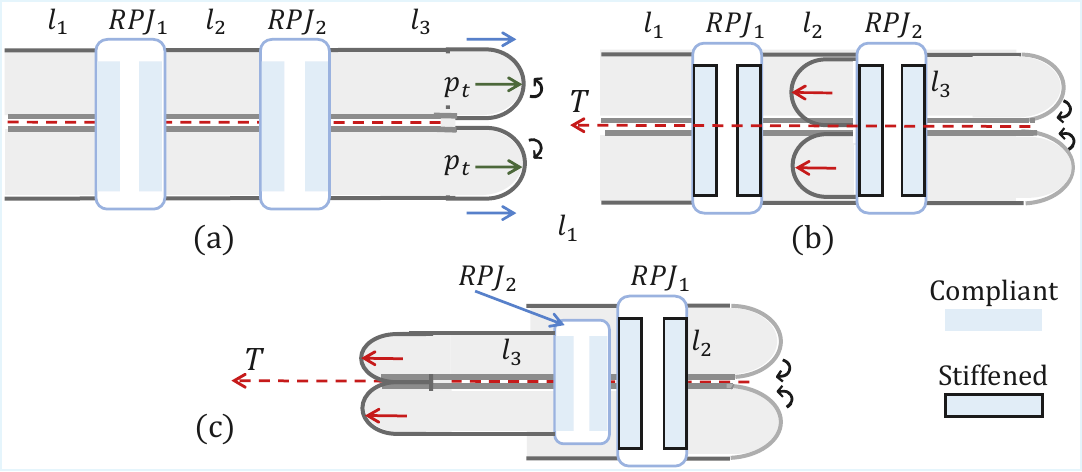}
\caption{Illustration of the RPJ-enabled cascading retraction sequence.}
\label{fig:retraction_mechanism}
\end{figure}

In addition to stiffness modulation during steering, the RPJ architecture enables sequential, or \textit{cascading}, retraction of the vine robot, as illustrated in Fig.~\ref{fig:retraction_mechanism}(a)--(c). By selectively pressurizing an upstream RPJ, a temporary stiff boundary is established, stabilizing the proximal structure while a downstream segment is withdrawn. This process is applied iteratively, resulting in controlled, segment-wise retraction.

During retraction, the robot is pulled inward from the tip, requiring inversion of the everted body. For a segment of length $l_s$, the applied tension $T$ must overcome the net resisting force $F_{\mathrm{net}}$ acting at the eversion front, as in (\ref{eq:pull_force1}):
\begin{equation}
T \ge F_{\mathrm{net}}
\label{eq:pull_force1}
\end{equation}

This resistance consists of the pressure-induced axial force due to the internal trunk pressure $p_t$, in addition to structural and frictional contributions associated with inversion of the LDPE body and embedded RPJ fabrics. Since the pressure-induced component is given by $p_t A$, the remaining effects are collectively represented by $F_{\mathrm{res}}$. The retraction condition can therefore be written as in (\ref{eq:pull_force}), where $A$ is the effective cross-sectional area at the eversion front:
\begin{equation}
T \ge p_t A + F_{\mathrm{res}}
\label{eq:pull_force}
\end{equation}

Importantly, the RPJs are depressurized prior to retraction and therefore do not generate active resisting forces. Their contribution is limited to passive structural effects already captured within $F_{\mathrm{res}}$. Accordingly, (\ref{eq:pull_force}) provides a practical estimate of the minimum tension required for retraction of an RPJ-based vine robot.

\subsection{RPJ Node Fabrication and Integration}
\label{subsec:prototyping}

\begin{figure}[t!]
    \centering
    \includegraphics[width=\linewidth]{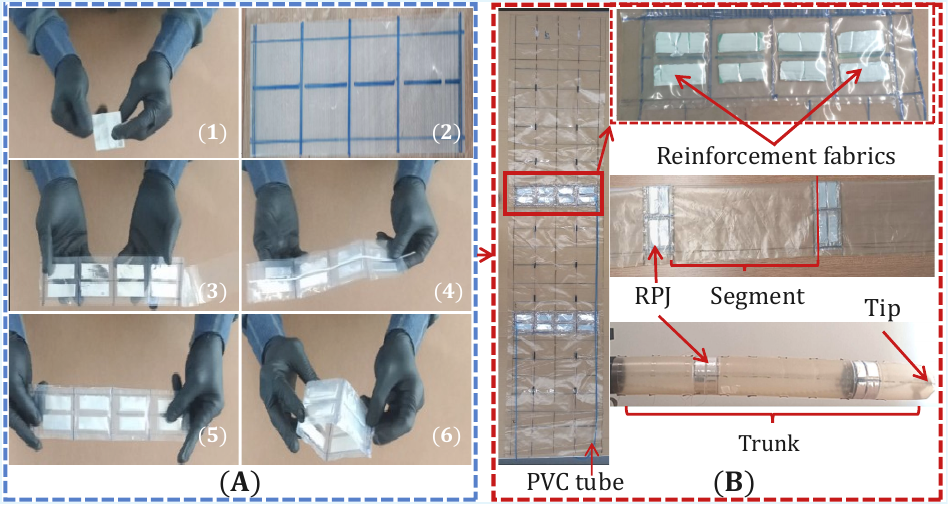}
    \caption{Fabrication sequence of the RPJ-based vine robot. (A) Formation of the RPJ node. (B) Integration of $3$~mm PVC tubing for pneumatic supply, followed by incorporation of tendons and mechanical stoppers (\textit{left}). \textit{Right}: The structure is longitudinally sealed and closed at one end, forming a tubular trunk comprising of RPJ--separated segments (\textit{middle and bottom}).}
    \label{fig:vinearm_design}
\end{figure}

The fabrication of the RPJ-based vine robot is guided by three objectives: (i) maintaining structural continuity for uninterrupted eversion, (ii) enabling independent pressure regulation between the trunk and RPJ modules, and (iii) preserving lightweight compliance while providing localized reinforcement. The workflow, illustrated in Fig.~\ref{fig:vinearm_design}A, follows:

\begin{enumerate}
    \item An extensible material is cut to the prescribed chamber dimensions. A $3$~mm thick BMPAPER cellulose rolling fabric is selected to balance flexibility with structural support during eversion.
    
    \item Two LDPE sheets ($270 \times 70$~mm) are prepared and marked to define four equal chambers, establishing the RPJ geometry relative to the trunk diameter.
    
    \item The fabrics from Step~1) are symmetrically placed within each chamber, $3$~mm from the upper and lower boundaries, and secured using double-sided adhesive.
    
    \item A removable barrier is inserted between the fabric layers to define the internal flow path. The exposed edges are heat-sealed to form the chamber boundaries.
    
    \item The barrier is removed, and the remaining edges are selectively heat-sealed to complete the chamber structure.
    
    \item A node is formed by sealing the structure into a closed ring, while RPJs are created by sealing at discrete axial locations along the robot body (Fig.~\ref{fig:vinearm_design}(B)).
\end{enumerate}

Fabrication fidelity is critical to achieving predictable RPJ behavior. Deviations in chamber symmetry result in non-uniform pressure distribution and asymmetric force generation, introducing unintended bending moments and reducing stiffness controllability. Similarly, variations in seam alignment alter the effective chamber area $A_c$ and deformation profile, directly affecting the pressure--force relationship.

For the characterization of our system and comparative evaluation, we fabricated four variants of vine robots: (i) a baseline trunk without RPJ structures; (ii) a trunk with RPJ chambers but without fabric reinforcement; (iii) the fully integrated RPJ configuration; and (iv) a vine robot incorporating layer-jamming. The latter serves as a benchmark against established localized stiffness strategies, particularly layer jamming~\cite{do2024stiffness, li2025enhanced, exarchos2022task}. All prototypes share identical global dimensions, with an inflated diameter of $85$~mm and a length of $1$~m.

\section{System Characterization}
\label{sec:characterization}

In this section, we characterize the operational limits of the proposed RPJ-based vine robotic system and perform a comparative evaluation against a vine robot incorporating paper layer-jamming mechanisms. Together, these results establish baseline performance metrics for subsequent demonstrations.

\subsection{Pressure Requirement for Robot Growth}
\label{subsec:pressure_requirements}

To quantify the influence of the RPJ on vine robot eversion, we evaluated the growth pressure across the first three structural configurations highlighted in Section.~\ref{subsec:prototyping}. The experimental setup comprised a precision pressure regulator (CROX AR2000 1/4 Regulator + Pressure Gauge) for controlled pressure input and a synchronized pressure sensor for measurement. Two key pressure metrics were recorded: the growth initiation pressure $P_{\text{init}}$, defined as the onset of tip eversion, and the steady growth pressure $P_{\text{grow}}$, corresponding to continuous extension at a constant velocity of $0.05$~m/s over a length of $1$~m. For each case, compressed air was supplied through the base station while the internal pressure was gradually increased until sustained eversion was achieved. 

\begin{figure}[h!]
    \centering
    \includegraphics[width=\linewidth]{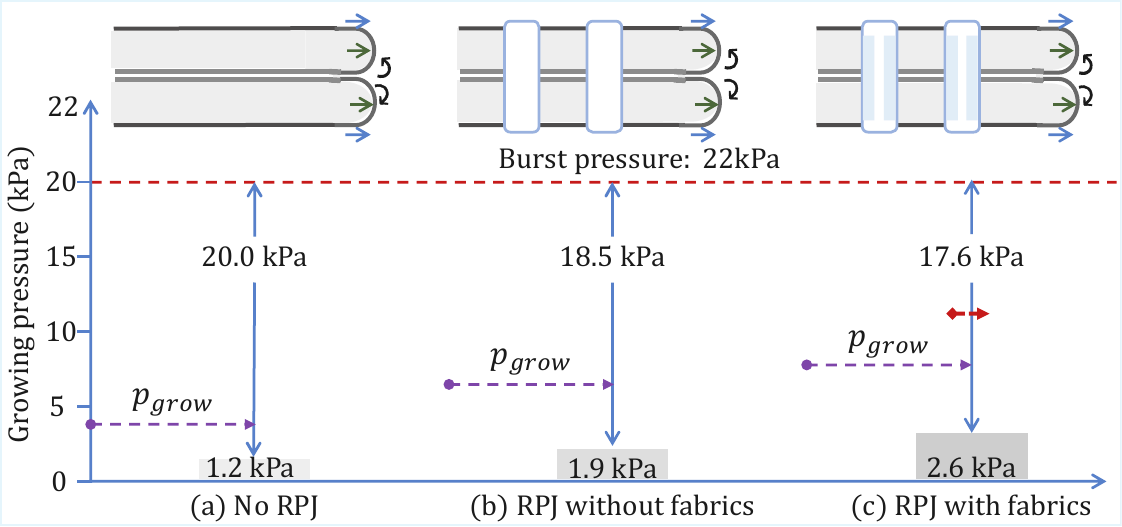}
   \caption{Minimum pressure required for growth initiation ($P_{\text{init}}$) and steady eversion ($P_{\text{grow}}$) across the three configurations (a)--(c).}
   \label{fig:pressure_requirments}
\end{figure}

As shown in Fig.~\ref{fig:pressure_requirments}, the baseline trunk exhibits the lowest growth initiation pressure ($P_{\text{init}} = 1.2$~kPa). Introducing RPJ chambers increases this requirement to $1.9$~kPa due to the added internal membrane constraints. With the inclusion of reinforcement fabrics, $P_{\text{init}}$ further increases to $2.6$~kPa, as the fabrics impose localized circumferential constraints that restrict radial expansion during pressurization. Notably, the inclusion of reinforcement fabrics reduces the margin between growth initiation and burst pressure by approximately $15\%$, indicating only a moderate impact on eversion capability.

The steady growth pressures $P_{\text{grow}}$ for the three configurations are $4.2$~kPa, $6.0$~kPa, and $6.8$~kPa, respectively. We realized that the RPJ-based trunk exhibits behavior comparable to the baseline during steady eversion, with deviations arising primarily when an RPJ module passes through the eversion front. At this stage, the chamber structure and embedded fabrics must invert along with the trunk material, resulting in a transient pressure increase. Specifically, $P_{\text{grow}}$ rises from $6.8$~kPa to approximately $12$~kPa (red arrow) to facilitate smoother inversion. While eversion remains feasible at the nominal pressure ($6.8$~kPa), this temporary increase improves inversion stability at the expense of a slight increase in growth time. Meanwhile, once the RPJ module is fully everted, the system returns to nominal steady-state growth conditions.

\subsection{RPJ Operating Pressure}
\label{subsec:RPJ_failure_test}

To establish a safe and repeatable operating range for the RPJ, we conducted an empirical destructive pressure test. The chamber pressure was incrementally increased from $1$~kPa in steps of $1$~kPa until structural failure occurred. The standalone RPJ node (Fig.~\ref{fig:contact_force_experimentsetup}(A)) exhibited the onset of irreversible bulging at approximately $19.5$~kPa, followed by rupture at $21.4$~kPa. When integrated with a pressurized trunk ($p_t = 12$~kPa) (Fig.~\ref{fig:contact_force_experimentsetup}(B)), a confinement effect was observed, resulting in an increased burst pressure of approximately $23$~kPa. Based on these results, we selected an operating pressure of $15$~kPa to ensure operation within the elastic deformation regime while providing sufficient stiffness modulation.

The pre-failure bulging behavior is attributed to the viscoelastic response of the LDPE material. This response is influenced by the film thickness ($75~\mu$m) and density ($0.915$~g/cm$^3$), which promote localized yielding under sustained internal pressure. It is important to note that this study employs positive pressure actuation as an initial proof-of-concept, driven by experimental constraints and system availability. However, the RPJ architecture is inherently compatible with vacuum-based actuation, which may offer alternative stiffness modulation characteristics. A systematic investigation of material-dependent behavior and vacuum-induced stiffness can be explored in future work.

\subsection{Distributed Contact Force Measurement}
\label{subsec:contact_force}

\begin{figure}[h!] 
\centering \includegraphics[width=\linewidth]{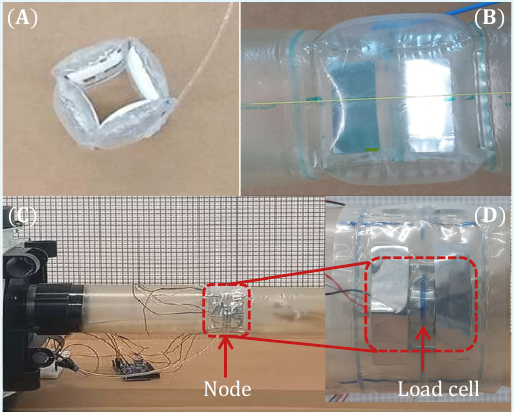} \caption{(A) Standalone RPJ node. (B) RPJ node integrated with the vine trunk. (C) Side view showing a load cell positioned between the RPJ node and the inner wall of the trunk. (D) Zoomed view illustrating local trunk compression during RPJ actuation.}
\label{fig:contact_force_experimentsetup} 
\end{figure}

To characterize the contact mechanics between the RPJ and the vine trunk, we mounted a single RPJ onto the trunk (Fig.~\ref{fig:contact_force_experimentsetup}(C)). We then implemented a distributed force measurement setup using four resistive load cells (HX-711 pressure module; 50~kg capacity and $0.2\%$ full-scale accuracy), circumferentially arranged between the trunk and the RPJ. Each sensor was aligned with the center of a chamber interface to capture the local normal force transmitted by each chamber. The four-sensor configuration directly reflects the four-chamber architecture of the RPJ, enabling spatially resolved measurement of circumferential contact forces.

The RPJ chamber pressure $p_j$ was incrementally increased from $1.5$ to $15$~kPa via a $4$~mm pneumatic supply line. We then recorded the resulting force distribution over $n=3$ trials, with each load cell independently measuring the local normal force. This configuration provides a direct mapping between chamber pressure and distributed contact force.

\begin{figure}[h!] 
\centering \includegraphics[width=\linewidth]{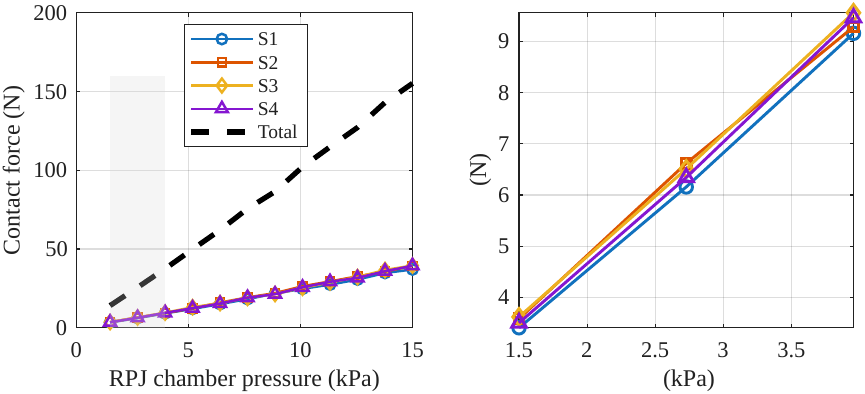} \caption{Distributed contact force as a function of RPJ pressure. (\textit{right}) Individual sensor responses show a near-synchronous, monotonic increase with pressure, while the total contact force (dashed line) follows a quasi-linear trend, reaching approximately $35$~N at $15$~kPa. The shaded region indicates the pressure range of minimal inter-sensor variation. (\textit{left}) Zoomed view of high-pressure convergence, with inter-sensor variation.}
\label{fig:contact_force} \end{figure}

As shown in Fig.~\ref{fig:contact_force}, all sensors exhibit a near-synchronous, monotonic increase in contact force with increasing $p_j$. At the maximum pressure ($15$~kPa), individual sensor readings converge within a narrow band (Fig.~\ref{fig:contact_force}~\textit{left}), with variations limited to $5$--$8\%$ (Fig.~\ref{fig:contact_force}~\textit{right}). These deviations may likely be attributed to minor fabrication-induced geometric asymmetries in the RPJ node. Additionaly, the total contact force increases quasi-linearly across the operating range, reaching approximately $35$~N at $15$~kPa. This behavior supports the assumption in Section~\ref{subsec:analytical_model} of an approximately uniform contact pressure distribution over each chamber footprint.

\subsection{Tip Deflection and Load-Bearing Analysis}
\label{subsec:RPJ_stiffness}

To evaluate the load-bearing capability of our proposed robotic system, we conducted a transverse tip deflection experiment under incremental external loading. The objective is to quantify the maximum supported load prior to excessive deformation and to assess the stiffness contribution of the RPJ relative to the baseline vine structure. 

\begin{figure}[t!]
    \centering
    \includegraphics[width=\linewidth]{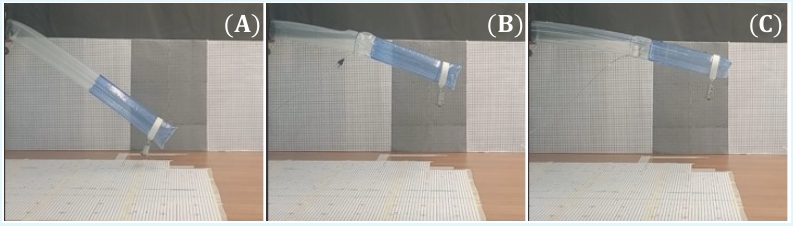}
    \caption{Final state of tip–deflection under a $200$~g load for: (A) baseline vine, (B) vine with unreinforced RPJ, and (C) vine with reinforced RPJ.}
    \label{fig:RPJ_trunktip_deflection}
\end{figure}

The experimental setup consists of a $1$~m long vine robot with a single RPJ module ($70$~mm length) positioned at mid-span, resulting in two distal segments of approximately $0.465$~m each. The robot is clamped at the base in its uninflated state. When inflated, the structure assumes a cantilever boundary condition. To measure vertical deflection, we  attached a loading hook $50$~mm from the tip, and a reference marker is used against a $1$~mm resolution grid (Fig.~\ref{fig:RPJ_trunktip_deflection}(A)--(C)). We set the initial tip height to $0.55$~m to ensure unconstrained deflection throughout the loading range.

\begin{figure}[t!]
    \centering
    \includegraphics[width=\linewidth]{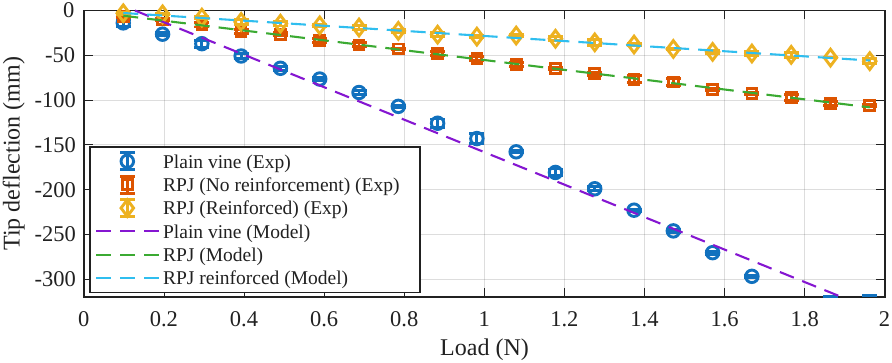}
    \caption{Load-deflection response of the vine robot for the three configurations in Fig.~\ref{fig:RPJ_trunktip_deflection} (A-C). Markers denote experimentally measured deflection (mean $\pm$ standard deviation, $n=3$), and dashed lines represent linear model fits.}
    \label{fig:RPJ_trunktip_deflection_plots}
\end{figure}

As defined in Section~\ref{subsec:pressure_requirements}, we evaluated three configurations with trunk and RPJ pressures maintained at $p_t = 12$~kPa and $p_j = 15$~kPa, respectively. Specifically, we applied loads in increments of $10$~g up to $200$~g, and recorded the corresponding tip deflection. Each measurement was repeated $n=3$ times and averaged. Fig.~\ref{fig:RPJ_trunktip_deflection_plots} shows the load--deflection response for all configurations. The baseline vine exhibits the largest deformation, reaching tip deflections of approximately $0.25$--$0.30$~m at maximum load, reflecting its low effective bending stiffness despite internal pressurization. The introduction of the RPJ reduces deflection by approximately $40$--$60\%$ across the loading range, while the reinforced RPJ achieves a further reduction of up to $70$--$80\%$, limiting peak deflection to below $0.10$~m. These results indicate stiffness increase of about $3$--$4\times$ for the reinforced configuration relative to the baseline. 

\subsection{Comparison with Layer Jamming Stiffness Modulation}
\label{subsec:layer_jamming_comparison}
\begin{figure*}[t!]
    \centering
    \includegraphics[width=\linewidth]{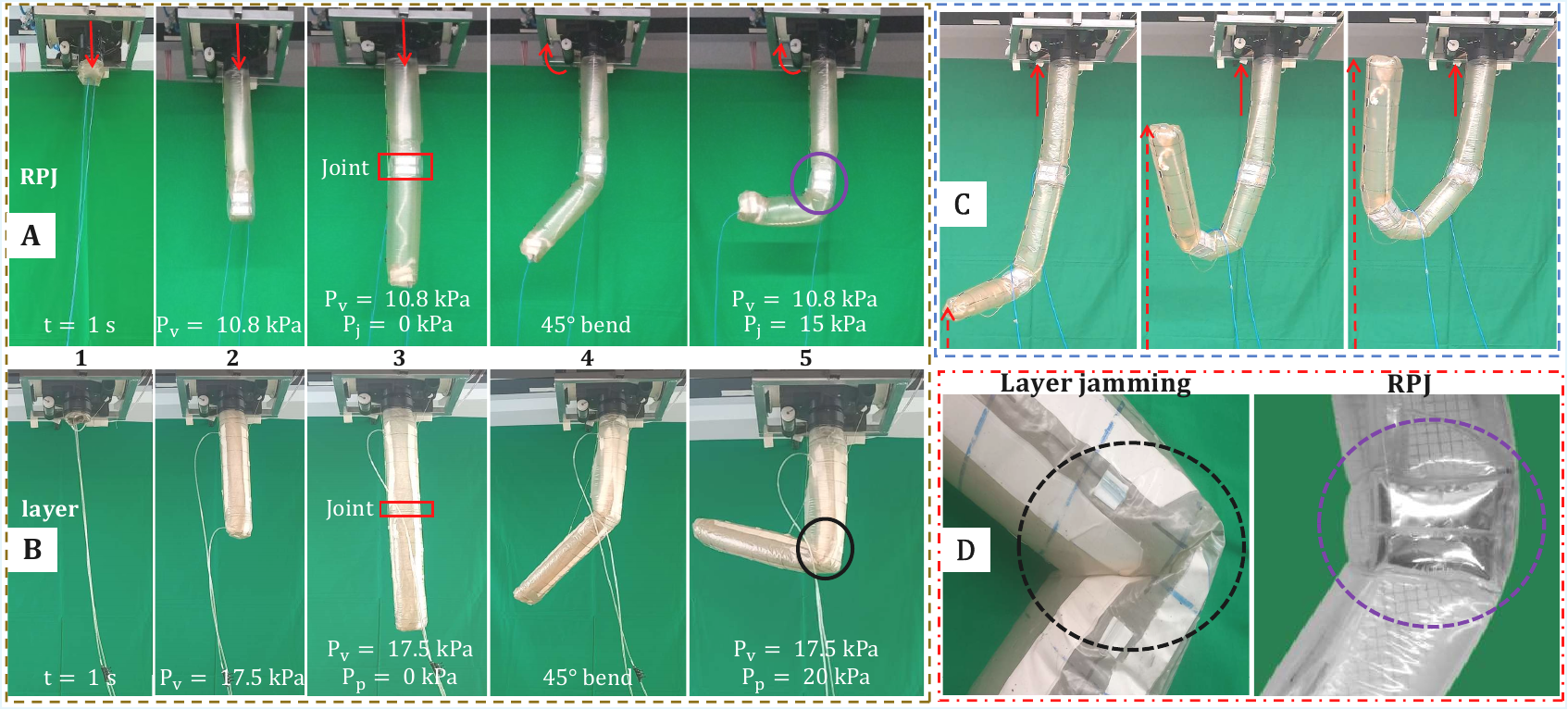}
    \caption{Comparison of RPJ-based (A) and layer-jamming arms (B) across identical operational stages: (1) onset of eversion ($t{=}1$~s), (2) completion of the first link prior to stiffening, (3) full two-link growth including the joints, (4) tendon-driven $45^{\circ}$ steering, and (5) high-curvature bending. (C) nature of buckling/material folding at the joints during bends, and (D) adaptive morphologies under continuous tendon pulling in the RPJ-based vine robot.}
    \label{fig:Comparison_demos}
\end{figure*}

We performed gravity-opposed eversion and planar bending experiments (Fig.~\ref{fig:Comparison_demos}A, B) to benchmark the proposed RPJ-based vine robot (Case~A) against a conventional layer-jamming architecture (Case~B)~\cite{do2024stiffness, li2025enhanced}. Three performance metrics are considered: (i) growth response time, (ii) localized bending stiffness, and (iii) maximum achievable curvature under comparable actuation conditions. Both systems are implemented as two-link robotic arms with a mid-span stiffening element, realized as an active RPJ in Case~A and a full-segment paper-jamming pouch in Case~B. Geometric parameters are consistent with Section~\ref{subsec:prototyping}.

Actuation is provided by Faulhaber 3242X012BX4--3692 brushless DC motors with IER3-10000L encoders and $20{:}1$ planetary gearboxes, driven by MCBL3006~S~RS drivers. Motor commands and encoder feedback are handled via a Dynomotion Kanalog--KFlop real-time controller at $100$~Hz. 

\begin{figure}[h!]
    \centering
    \includegraphics[width=\linewidth]{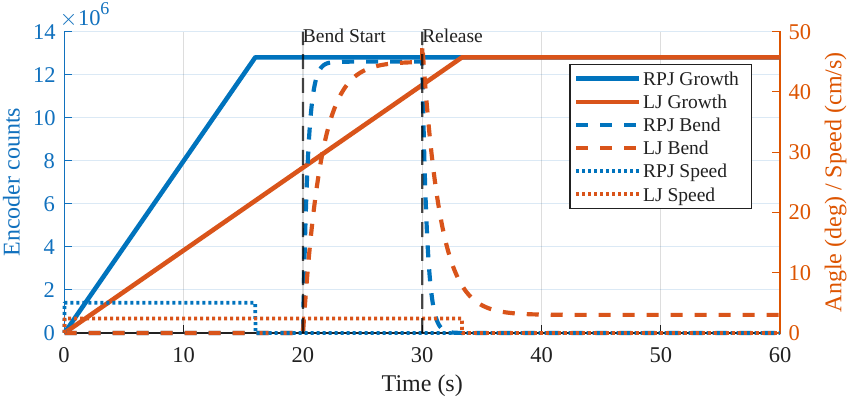}
    \caption{Quantitative comparison of growth, bending response, and actuation dynamics between the proposed RPJ-based robot and a layer-jamming system. The RPJ design exhibits faster growth, rapid curvature response, and stable recovery, while the layer-jamming system shows slower dynamics and residual deformation. Phase markers indicate bending onset and release.}
    \label{fig:comparison_data}
\end{figure}

During eversion, the RPJ-based robot achieved an initial growth speed of approximately $0.05$~m/s at $10.8$~kPa (Fig.~\ref{fig:Comparison_demos}, Case~A1--3), compared to $2.4 \times 10^{-4}$~m/s for the layer-jamming system under the same pressure (Fig.~\ref{fig:Comparison_demos}, Case~B1--3). The reduced operating pressure ($10.8$~kPa versus $12$~kPa in Section~\ref{subsec:pressure_requirements}) reflects lower gravitational loading during early-stage eversion, where only a short segment is exposed. For a deployed length of $\sim 1$~m, the RPJ completed eversion in approximately $18$~s, consistent with the imposed feed rate. In contrast, the layer-jamming arm required increased pressure (up to $17.5$~kPa) to sustain growth and reached full extension in approximately $34$~s. The RPJ system maintained stable growth throughout, whereas the layer-jamming arm exhibited intermittent stalls due to increased distal resistance.

Furthermore, to evaluate localized bending stiffness, the RPJ was pressurized to $p_j = 15$~kPa (Case~A4) and the jamming pouch to $p_p = 20$~kPa~\cite{do2024stiffness} (Case~B4), followed by a tendon-driven $45^{\circ}$ deflection. Both systems maintained structural integrity without global buckling. Under continued planar bending (Case~A5 and B5), distinct deformation behaviors emerge. The RPJ-based robot achieves smooth curvature profiles up to approximately $90^{\circ}$ (Fig.~\ref{fig:Comparison_demos}, Case~A5). With continued tendon actuation, as illustrated in Fig.~\ref{fig:Comparison_demos}(C), the robot transitions to a configuration that distributes curvature along the trunk, rather than undergoing localized structural collapse. This response reflects the RPJ’s ability to decouple localized stiffness from global compliance. In contrast, the layer-jamming arm exhibits progressive axial contraction with increasing curvature, leading to telescoping of the structure and limiting the achievable bending angle. The distinct bending behaviors of both joints are illustrated in Fig.~\ref{fig:Comparison_demos}(D).

Generally, the faster eversion and adaptive morphology of the RPJ-based vine robot stem from its localized, low-mass stiffening mechanism, which introduces negligible resistance to material transport and preserves continuous tip growth. In contrast, layer jamming imposes distributed axial drag and distal mass loading, increasing the pressure required for eversion and constraining morphological adaptability. While layer jamming provides stronger whole-segment stiffening, which effectively mitigates buckling at the base, comparable proximal rigidity in the RPJ system can be achieved through strategic placement of one or more joints near the outlet. Table~\ref{tab:quantitative comparison} and Fig.~\ref{fig:comparison_data} summarize the quantitative comparison between both systems

\begin{table}[h]
\centering
\small
\caption{Comparison of RPJ and Layer-Jamming Performance}
\begin{tabular}{lccc}
\hline
\textbf{Metric} & \textbf{RPJ} & \textbf{LJ} & \textbf{Improvement} \\
\hline
Growth time (s) & 18.0 & 34.0 & $\approx1.9\times$ faster \\
Steady growth speed (cm/s) & 5.0 & 2.4 & $\approx2.1\times$ faster \\
Peak curvature (deg) & $\geq 90$ & $\approx 100$ & -- \\
Rise time to full curvature (s) & 0.40 & 2.00 & $5.0\times$ faster \\
\hline
\end{tabular}
\label{tab:quantitative comparison}
\end{table}

\section{Experimental Validations}
\label{sec:demostrations_experiments}
To validate the results in Section~\ref{sec:characterization} and assess capabilities of the proposed system, we conducted task demonstrations on shape locking, cascading retraction, and payload integration.

\subsection{Shape Locking and Selective Stiffening in Free Space}
\label{subsec:navigation_through_obstacle_field}

In addition to the adaptive morphology shown in Fig.~\ref{fig:Comparison_demos}(C), where the robot is suspended from the ceiling, we demonstrate that the RPJ-based vine robot enables selective stiffening and shape locking while fully supporting its own weight in free space. For this experiment, the robot was extended to a length of $1.5$~m, comprising three RPJs (two actively engaged) and three links of lengths $0.30$, $0.50$, and $0.40$~m, respectively. We developed a compact base station that enables our robot to initiate growth at a height of $0.115$~m above ground level.

\begin{figure}[t!]
    \centering
    \includegraphics[width=\linewidth]{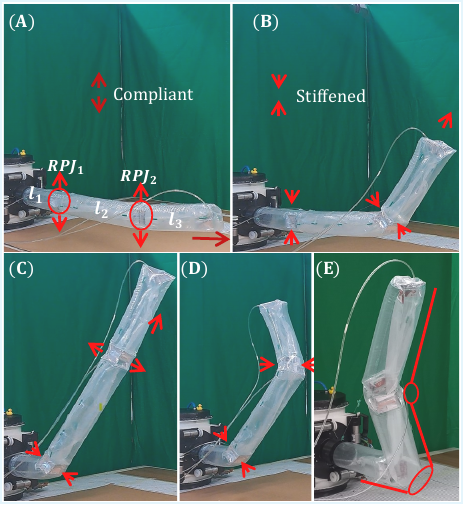}
    \caption{Annotated sequence of selective RPJ stiffening and shape locking in free space. (A) The robot grows to $1.5$~m, revealing two unactuated RPJs and three links ($l_1$, $l_2$, $l_3$). (B) Both RPJs are stiffened; the distal link ($l_3$) bends while the rest remains straight. (C) RPJ$_2$ is released while RPJ$_1$ remains stiff, enabling bending at Joint~1 and lifting of $l_2$ and $l_3$. (D) Both RPJs are re-stiffened, and $l_3$ is bent to form and lock an arm-like configuration. (E) Complex shapes are achieved through sequential joint locking and release.}
    \label{fig:shape_locking}
\end{figure}

Fig.~\ref{fig:shape_locking} (A)--(D) presents a time-lapse sequence of sequential RPJ activation and link bending. The target configuration is achieved through alternating deactivation and reactivation of RPJs, starting from the distal segment $l_3$ and progressing proximally. This spatial stiffness patterning enables independent joint actuation using a single tendon, as localized stiffening effectively isolates the mechanical response of each segment.

We observed an agreement between the model in Section~\ref{subsec:tendon_bending} and the physical operation of our robot. The seam-bonded RPJ architecture enables well-defined, discrete joint behavior while supporting multi-directional bending. During bending, deformation localizes at the RPJ interface, preserving the effective length of the upstream link while inducing minor, confined buckling at the seam between the RPJ and the downstream segment (Fig.~\ref{fig:shape_locking} (B)--(E)). This preserved and clear segmentation between links notably expands the achievable workspace of the robot, positioning the system for manipulation-oriented tasks such as object sorting and adaptive exploration in unconstrained environments. In addition, the joint chamber architecture promotes passive rotation around the joint chambers, as shown in Fig.~\ref{fig:shape_locking} (E).

\subsection{Cascading Retraction}
\label{subsec:cascading_retraction_demo}

\begin{figure}[t!]
    \centering
    \includegraphics[width=\linewidth]{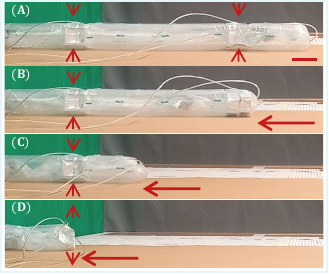}
    \caption{Annotated sequence of RPJ-enabled \textit{cascading} retraction. (A) The robot is fully everted to a length of $1.5$~m. (B) Both RPJs are pressurized to establish stiff boundaries of $15$ kPa, and segment $l_3$ is retracted up to $\mathrm{RPJ}_2$. (C) After depressurizing $\mathrm{RPJ}_2$, the proximal segment and $l_2$ are retracted up to $\mathrm{RPJ}_1$. (D) Finally, $\mathrm{RPJ}_1$ is depressurized, completing retraction of the first two segments.}
    \label{fig:cascading_retraction}
\end{figure}

We demonstrate cascading retraction using a fully everted $1.5$~m robot supported by the environment. The retraction sequence is shown in Fig.~\ref{fig:cascading_retraction}(A)--(D). At each stage, the upstream RPJ is pressurized to $p_j = 15$~kPa, while the active trunk segment is depressurized to $6$~kPa (i.e., $0.5\,P_{\text{grow}}$), and the tail is pulled at a constant speed of $0.02$~m/s. As the internal pressure decreases, distal segments retract toward the base in a controlled manner. Because the RPJ modules remain pressurized, each reinforced node preserves local rigidity during retraction, preventing global collapse and promoting sequential folding of the trunk into the base. This behavior is consistent with the model presented in Section~\ref{subsec:cascading_retraction}. Compared with existing retraction strategies~\cite{coad2020retraction, kim2023self, jeong2020tip}, this approach requires no additional onboard mechanisms or external tools. Retraction is achieved solely through coordinated modulation of $p_t$ and $p_j$, enabling a simplified and fully integrated bidirectional growth--retraction process.

While cascading retraction can also be achieved in free space, our empirical observations indicate that stable operation without environmental support is limited to relatively short vine lengths on the order of $0.2$--$0.6$~m with a single mid-span RPJ. As the robot length increases, gravitational loading combined with low global bending stiffness leads to structural sagging, which disrupts load transmission along the trunk and degrades retraction controllability. Consequently, the sequential folding mechanism becomes less effective, and uncontrolled deformation may occur. Hence, for longer robots, either environmental support or additional strategically distributed RPJs are required to maintain stability and preserve controlled cascading behavior during free-space retraction.

\subsection{Payload Interaction}
\label{subsec:payload_interaction}

In Section~\ref{subsec:RPJ_stiffness}, we evaluated the static tip load-bearing capability of the RPJ-based robot. However, practical deployment requires simultaneous load support and controlled manipulation; capabilities that remain challenging for vine robots due to their compliant structure and limited axial stiffness, particularly during unsupported (free-space) eversion.

\begin{figure}[t!]
    \centering
    \includegraphics[width=\linewidth]{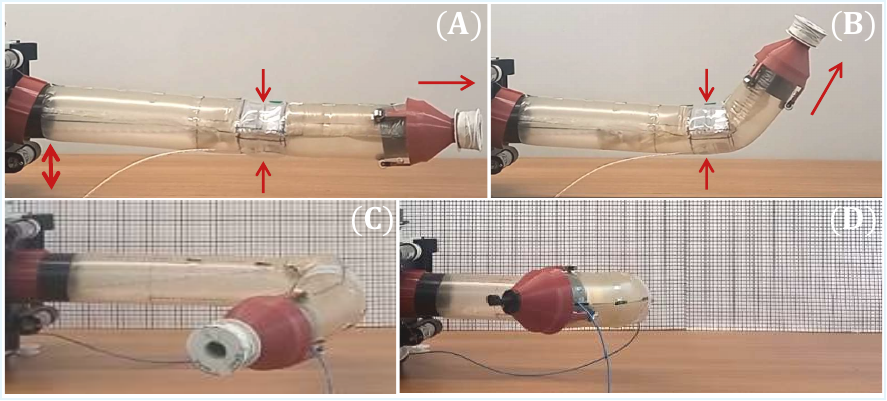}
    \caption{Payload-bearing demonstration of the RPJ-based vine arm in free space (11.5 cm above ground). (A) The robot everts with a $202$ tip load (end effector + 100 g mass). (B) Robot performs a $45^{\circ}$ upward bend while sustaining the full payload (C) Repeating the bend with only the end effector achieves smoother curvature.}
    \label{fig:Load_bearing experiment}
\end{figure}

To investigate this, we evaluated the RPJ-based vine robot under payload in free space (Fig.~\ref{fig:Load_bearing experiment}). A compact tip-mounted module was developed to enable payload integration without disrupting eversion. The module consists of (i) an outer cap for payload attachment and (ii) an internal magnetic--bearing interface that provides secure mechanical coupling to the vine body. A precision roller bearing ensures low-friction material flow during eversion, while integrated cutouts allow pneumatic lines to pass through without altering the growth path. The assembled end effector has a mass of $102$~g.

An additional $100$~g mass was attached, resulting in a total payload of $202$~g. The robot was deployed at a trunk pressure of $p_t = 12$~kPa with RPJ actuation at $p_j = 15$~kPa, achieving full $1$~m eversion (Fig.~\ref{fig:Load_bearing experiment}(A)). During growth, intermittent tip relaxation and slight deviation from linearity were observed as the length increased. These effects arise from payload-induced bending moments combined with increasing gravitational loading along the cantilevered structure, which locally exceed the pressure-induced axial stiffness during dynamic eversion. Similar to the free space limitation in Section~\ref{subsec:cascading_retraction_demo}, this behavior does not reflect a limitation of the RPJ mechanism itself. Rather, it highlights the inherent challenge of maintaining structural support and continuous growth in fully unsupported configurations. When environmental support is available, higher payloads can be accommodated due to distributed load sharing along the robot body.

We further demonstrate that tendon actuation can compensate for these effects by introducing counteracting moments. As shown in Fig.~\ref{fig:Load_bearing experiment}(B and C), tendon-assisted segments maintain a stable $45^{\circ}$ and about $95^{\circ}$ upward and sideways configurations respectively under load, effectively suppressing payload-induced sagging. In the absence of the additional $100$~g mass (Fig.~\ref{fig:Load_bearing experiment}(D)), the robot achieves the same deployment without observable relaxation under identical pressure conditions, confirming that the RPJ modules provide sufficient localized stiffness for lightweight end-effectors such as grippers, cutters, or embedded sensing units.

\section{Conclusion}
\label{sec:discussion_conclusion}
This work presents the RPJ mechanism as a physically grounded approach to localized stiffness modulation in vine robots. Unlike strategies based on global pressurization or distributed reinforcement, the RPJ enables discrete, independently actuated stiffness nodes, allowing curvature generation and spatial regulation without compromising overall compliance. This facilitates piecewise deformation control while preserving the inherent adaptability of soft growing robots.

From the contact force experiments Section~\ref{subsec:contact_force}, the near-linear increase in force with chamber pressure translates directly into controllable normal reaction forces at the robot--environment interface. This establishes a predictable mapping between actuation input and local rigidity. Consequently, the RPJ enables piecewise curvature control, effectively discretizing the trunk (backbone) while retaining compliance---a capability that is difficult to achieve in continuum soft robots, where curvature arises from global actuation. This distinction is critical for directional growth and obstacle negotiation, as uncontrolled compliance can lead to shape drift or loss of steering authority. Local stiffness modulation therefore allows the robot to maintain intended trajectories while adapting to environmental constraints.

A key implication is the ability to sustain self-weight during reconfiguration in free space (Section~\ref{subsec:navigation_through_obstacle_field}), contrasting with tip-everting robots that typically rely on environmental support~\cite{coad2019vine, exarchos2022task, blumenschein2020design} or exhibit limited load-bearing capacity when extended horizontally~\cite{li2025mixedlayerjamming, feteih2025active}. Increased chamber pressure enhances resistance to bending and collapse, suggesting that RPJ nodes act as distributed stiffness regulators that mitigate gravitational deflection. This enables stable configurations without continuous environmental bracing, enabling deployment in scenarios where intermediate support is unavailable. Importantly, this is not achieved through rigidization of the entire body, but through selective stiffening, preserving the fundamental compliance required for safe interaction.

The RPJ also enables cascading retraction, where segments are sequentially depressurized and inverted (Section~\ref{subsec:cascading_retraction} and \ref{subsec:cascading_retraction_demo}). In contrast to retraction strategies requiring external tools or dedicated subsystems~\cite{coad2020retraction, jeong2020tip}, the cascading retraction introduces a form of structural reversibility, where the same actuation pathways used for growth can be repurposed for controlled withdrawal. The localized nature of stiffness modulation ensures that retraction can occur without global collapse, as adjacent segments can remain temporarily stabilized during the process. This has practical implications for deployment in constrained or hazardous environments, where retrieval of the robot is necessary but external intervention is limited. It also reduces system complexity by eliminating the need for additional retraction subsystems.

While the current study validates quasi-static behavior, dynamic effects such as transient pressure fluctuations, viscoelastic response of the material, and time-dependent relaxation were not explicitly modeled. These factors may influence the stability of shape locking and the precision of curvature control during rapid actuation. Incorporating dynamic models and closed-loop pressure regulation, potentially augmented by embedded sensing, would enable more accurate and robust control in practical deployments.

Lastly, the current implementation relies on distributed pneumatic tubing to supply air to each RPJ chamber. As the number of joints increases, this results in increased routing complexity and potential flow delays due to pressure losses along the tube. These effects can limit scalability and reduce responsiveness, particularly in long-length deployments. A promising direction to address this challenge is the integration of multiplexed pneumatic architectures or local passive valves~\cite{do2020dynamically} to reduce system overhead.

Future work will focus on dynamic stiffness control during growth, integration of embedded sensing for closed-loop adaptation, and scalable pneumatic architectures for multi-joint systems. Further refinement of the analytical model to incorporate nonlinear deformation and long-length stability under self-weight will support more robust design and deployment in real-world environments.


%

\ifCLASSOPTIONcaptionsoff
  \newpage
\fi



%

\bibliographystyle{IEEEtran}
\bibliography{IEEEabrv,Bibliography}

%








\end{document}